%% file: root.tex
\documentclass[letterpaper, 10 pt, conference]{ieeeconf}

\IEEEoverridecommandlockouts                              

\usepackage{myStyle}

\begin{document}

\input{tex/title_authors}

\input{tex/abstract}


\input{tex/intro}

\input{tex/previous_work_new}

\input{tex/contributions.tex}

\input{tex/preliminaries}

\input{tex/proposed_approach}

\input{tex/experiments}

\input{tex/conclusions}


\bibliographystyle{IEEEtran}
\bibliography{IEEEabrv,cogro_mppi_bibtex}

\end{document}

%% file: tex/title_authors.tex
\title{\LARGE \bf
Multi-Agent Path Integral Control for\\Interaction-Aware Motion Planning in Urban Canals
}

\author{Lucas Streichenberg$^{1,2,*}$, Elia Trevisan$^{1,*}$, Jen Jen Chung$^{2,3}$, Roland Siegwart$^2$ and Javier Alonso-Mora$^1$ 
\thanks{This research is supported by the project ``Sustainable Transportation and Logistics over Water: Electrification, Automation and Optimization (TRiLOGy)'' of the Netherlands Organization for Scientific Research (NWO), domain Science (ENW), and the Amsterdam Institute for Advanced Metropolitan Solutions (AMS) in the Netherlands.}
\thanks{$^*$These authors have contributed equally.}
\thanks{$^{1}$Cognitive Robotics Department,
        TU Delft, 
        {\tt\small \{e.trevisan, j.alonsomora\}@tudelft.nl}}
\thanks{$^{2}$Autonomous Systems Lab, ETH Zurich
        {\tt\small \{stlucas, rsiegwart\}@ethz.ch}}
\thanks{$^{3}$School of ITEE, The University of Queensland
        {\tt\small jenjen.chung@uq.edu.au}}}

\maketitle
\thispagestyle{empty}
\pagestyle{empty}

%% file: tex/abstract.tex
\begin{abstract}
Autonomous vehicles that operate in urban environments shall comply with existing rules and reason about the interactions with other decision-making agents.  
In this paper, we introduce a decentralized and communication-free interaction-aware motion planner and apply it to Autonomous Surface Vessels (ASVs) in urban canals.
We build upon a sampling-based method, namely Model Predictive Path Integral control (MPPI), and employ it to, in each time instance, compute both a collision-free trajectory for the vehicle and a prediction of other agents' trajectories, thus modeling interactions. To  improve the method's efficiency in multi-agent scenarios, we introduce a two-stage sample evaluation strategy and define an appropriate cost function to achieve rule compliance. We evaluate this decentralized approach in simulations with multiple vessels in real scenarios extracted from Amsterdam's canals, showing superior performance than a state-of-the-art trajectory optimization framework and robustness when encountering different types of agents.
\end{abstract}


%% file: tex/intro.tex
\section{Introduction}
With rising population density, cities are forced to enhance their mobility and transportation strategies. The City of Amsterdam aims to reduce the load on road infrastructure by transporting goods and people on the urban waterways \cite{gemeente_amsterdam_uitvoeringsplan_2020}. This presents a great opportunity to operate Autonomous Surface Vessels (ASVs) such as Roboat \cite{leoni_roboat_2022} in urban canals. 
However, this is a very technically challenging task due to the complex and dynamic nature of the environment. Narrow canals, complex dynamics, static obstacles, and human-piloted vessels must be dealt with while obeying existing canal regulations \cite{koninkrijksrelaties_binnenvaartpolitiereglement_2017}. 
Model Predictive Path Integral Control (MPPI)\cite{Williams2015} offers a parallelizable sampling-based framework for solving motion planning tasks with such complex dynamics and discontinuous costs as those exhibited in our domain.
Unlike methods based on constrained optimization, which need to rely on convex approximations of the free space and on inflating the ego and obstacle agents into ellipsoidal shapes for collision avoidance~\cite{de_vries_regulations_2022}, MPPI can account for the exact and potentially non-convex shape of both static and dynamic obstacles. This promises to be a significant advantage in tight interaction-rich environments.
Still, another important aspect of achieving safe and efficient navigation in crowded spaces is accounting for cooperation~\cite{trautman_robot_2013}.
For these reasons, we propose a method to decentralize MPPI in order to navigate among non-communicating agents while providing interaction awareness and generating cooperative motion plans.
We introduce awareness of navigation rules through discontinuous costs.
Moreover, the proposed method can run in real-time thanks to our two-stage sample evaluation strategy and CPU parallelization.


\begin{figure}[t!]
    \centering
    \includegraphics[width=0.4\textwidth]{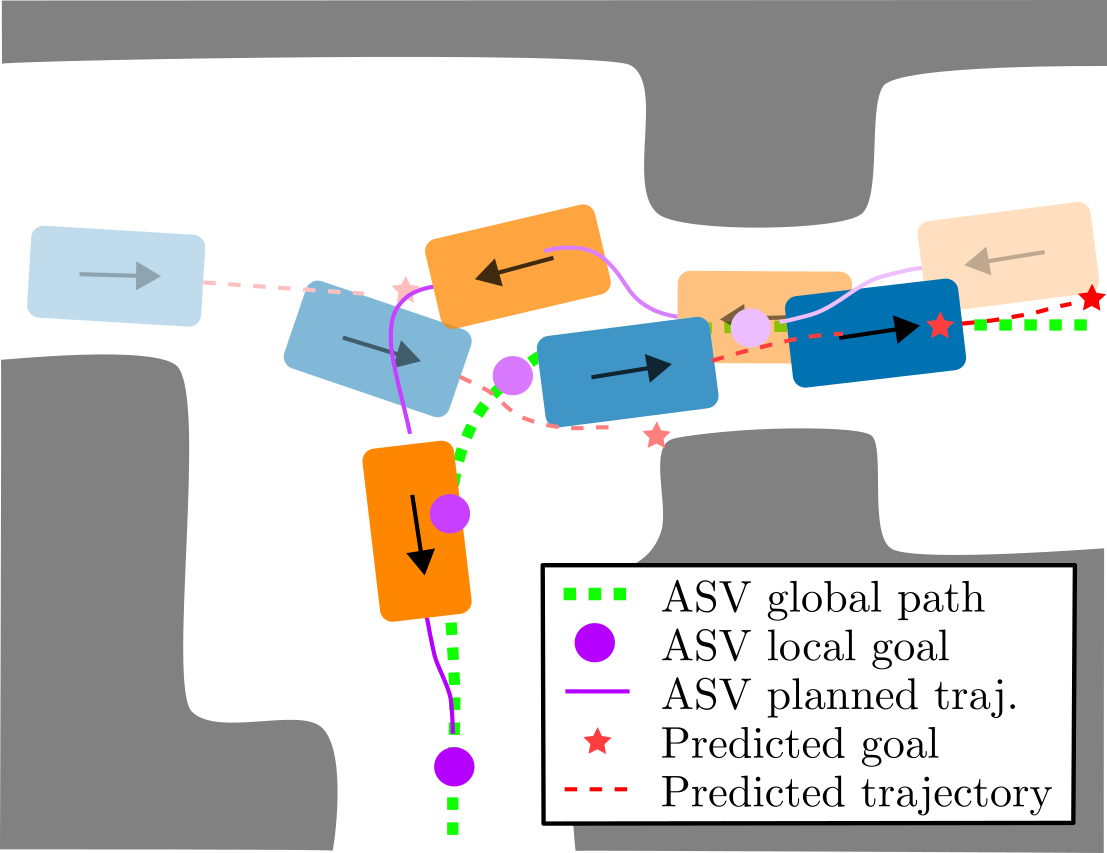}
    \caption{An ASV running our method (\textcolor{orange}{orange}) encounters a non-communicating vessel (\textcolor{blue}{blue}). At each time step, the ASV is given its current position and a global path (\textcolor{limegreen}{dashed green line}). Based on this, the ASV sets a local goal (\textcolor{purple}{purple circle}) in front of itself on the global path. The ASV is also given the position and velocity of the non-communicating vessel. With these, the ASV estimates the local goal of the non-communicating vessel (\textcolor{red}{red star}) assuming constant velocity. Then, the ASV plans input sequences over a defined horizon resulting in trajectories for both vessels.} 
    \label{img:illustration}
\end{figure}

%% file: tex/previous_work_new.tex
\subsection{Related Work} \label{sec:prevwork}
Cooperative and interactive motion planning for robotics is a challenging problem with a vast literature \cite{schwarting_planning_2018}.

The most well-known examples for planning in dynamic environments are the Dynamic Window Approach (DWA)~\cite{fox_dynamic_1997}, Reciprocal Velocity Obstacle (RVO) \cite{van_den_berg_reciprocal_2008}, its extension Optimal Reciprocal Collision Avoidance (ORCA)~\cite{van_den_berg_reciprocal_2011}, and Artificial Potential Fields (APF)~\cite{svenstrup_trajectory_2010, ji_path_2017}.
While these methods are highly efficient, they often lead to reactive behaviors. 

Model-free reinforcement learning algorithms have been successfully trained in simulation with hand-crafted reward functions to navigate among human crowds~\cite{chen_decentralized_2017, chen_socially_2017}, but generalization and collision avoidance are not guaranteed.

Game theoretic approaches have been implemented in the context of autonomous cars to perform lane changes~\cite{bouton_reinforcement_2020}, merging~\cite{garzon_game_2019} and to solve unsignalized intersections~\cite{tian_game-theoretic_2022}. However, they rely on a coarse discretization of the action space and do not scale well with the number of agents.

Model Predictive Control (MPC) based on trajectory optimization is a popular approach when it comes to local motion planning. In a multi-agent setting, however, the actions of the other agents are required to proceed with the motion planning of the ego-agent. In \cite{luis_online_2020} a distributed MPC was developed for motion planning with multiple drones relying on ideal communication. A way to avoid communication is to estimate each agent's state and predict their future motion using, for example, a constant velocity model~\cite{de_vries_regulations_2022, kamel_robust_2017} or learning-based techniques~\cite{zhu_learning_2021}.
However, as long as we first predict and then plan, large parts of the state-space can be perceived as unsafe by the motion planner~\cite{trautman_unfreezing_2010}.
This can be overcome by modeling interaction, such that the ego-agent is aware that its actions can influence the actions of the other agents around it. Unfortunately, accounting for such a model while planning with constrained optimization techniques can become computationally expensive~\cite{sadigh_planning_2016}.

In contrast to optimization-based methods, MPPI solves for the best control trajectory at each step by forward simulating the behavior of the full system. To achieve this, MPPI uses a parallelizable sampling-based framework~\cite{kappen_path_2005} to rollout simulations, allowing it to find an approximate solution to non-linear, non-convex, discontinuous Stochastic Optimal Control (SOC) problems~\cite{Williams2015,williams_information-theoretic_2018}. Compared to other SOC methods such as iterative Linear Quadratic Gaussian (iLQG)~\cite{li_iterative_2007} or Differential Dynamic Programming (DDP)~\cite{Theodorou2010a}, MPPI does not require linearization of the system dynamics or quadratic approximation of the cost function.
This makes MPPI particularly well-suited to our target task of ASV navigation in urban canals since the regulation-based interactions explicitly give rise to non-differentiable costs.
Moreover, MPPI's fast parallelizable computations could solve interaction-aware motion planning problems while still running in real-time.
When deployed to centralized multi-agent systems, however, the classic MPPI approach shows a significant increase in the number of samples required with an increasing number of agents~\cite{williams_model_2017}. 

%% file: tex/contributions.tex
\subsection{Contributions}\label{sec:contribution}

We propose an interaction-aware motion planning method based on MPPI which can generate cooperative plans in environments with non-communicating agents accounting for the full dynamics of the system and the exact shapes of the obstacles. To summarize our contributions:
\begin{itemize}
    \item We propose a decentralized architecture that can operate with limited or no communication under the assumption that other agents' states can be sensed exactly and that all the agents in the environment behave rationally.
    \item To reduce the number of samples required to plan for multi-agent systems, we propose a two-stage sample evaluation technique that improves sample efficiency.
    \item We formulate the objectives of the navigation task into an appropriate cost function to achieve rule compliance.
\end{itemize}
To demonstrate the performance of our method, we compare it to a state-of-the-art regulations-aware optimization-based MPC \cite{de_vries_regulations_2022} in several simulated experiments set up in real sections of Amsterdam's canals. The proposed decentralized MPPI is also compared to the centralized version in environments with crowds of up to four interacting vessels. Finally, we demonstrate the robustness of the algorithm in scenarios where a human-driven vessel does not behave rationally and provide insights into the computation times.








%% file: tex/preliminaries.tex
\section{Preliminaries}\label{sec:preliminaries}

We start by describing the basic MPPI algorithm. Then, since MPPI relies on a model of the system to perform the simulated rollouts, we also define the relevant dynamics that describe the behavior of our multi-ASV system.

\subsection{MPPI Algorithm}\label{sec:mppi_basic}
The presented work is based on the MPPI derivations by~\cite{williams_information-theoretic_2018}. With this method, we can solve SOC problems for discrete-time dynamical systems of the form,
\begin{equation}
    \mathbf{q}_{t+1} = \mathcal{F}(\mathbf{q}_t, \tilde{\mathbf{u}}_t),   \quad   \tilde{\mathbf{u}}_t \sim \mathcal{N}(\mathbf{u}_t,\mathbf{\Sigma}),
\end{equation}
with state $\mathbf{q}$, time step $t$, nonlinear state transition function $\mathcal{F}$ and noisy input $\tilde{\mathbf{u}}$ with variance $\mathbf{\Sigma}$ and mean $\mathbf{u}$, where $\mathbf{u}$ is the input we command to the system. The algorithm samples $K$ input sequences $\tilde{U}_k, \, k\in [1,K]$ from a distribution $\mathcal{N}(\mathbf{u}_t, \nu \mathbf{\Sigma})$ (with scaling parameter $\nu$)~\cite{williams_model_2019} and simulates them into $K$ state trajectories $Q_k$ over a horizon $T$ as,
\begin{equation}
Q_k = \big[\mathbf{q}_0, \, \mathcal{F}(\mathbf{q}_0, \tilde{\mathbf{u}}_{k,0}), \, \dots, \, \mathcal{F}(\mathbf{q}_{k, T-1}, \,\tilde{\mathbf{u}}_{k,T-1})\big].
\end{equation}
Each sample is rated by computing the total cost $S_k$, which includes a stage cost and a terminal cost.
\textit{Importance sampling} weights $w_k$ are then computed based on the cost of the sample $k$ minus the minimum sampled cost $S_{min}$ as,
\begin{equation}
    w_k = \frac{1}{\eta} \exp\biggl( \frac{-1}{\lambda}(S_k - S_{min})\biggr), \quad \sum_{k = 0}^{K-1} w_k  = 1, 
\end{equation}
with normalization factor $\eta$ and tuning parameter $\lambda$. The resulting control input sequence $U^*$, which approximates the optimal control input sequence, is computed with, 
\begin{equation}
    U^* =  \sum_{k = 0}^{K-1} w_k \tilde{U}_k. 
\end{equation}
Then, the first input $u_0^*$ of the computed sequence $U^*$ is applied to the system.

\subsection{Vessel State and Dynamics}\label{sec:state_dyn}
We define the multi-agent state similarly to~\cite{williams_model_2017}. The state of agent $i$ is defined as the concatenation of its position, heading, and associated velocities,
\begin{equation}\label{eq::dynamics}
    \mathbf{q}_i = \begin{bmatrix}  \mathbf{x}_i^\top &
                                    \mathbf{v}_i^\top \end{bmatrix}^\top
    = \begin{bmatrix} x_i &  y_i & \phi_i  & \dot{x}_i & \dot{y}_i & \dot{\phi}_i \end{bmatrix}^\top.
\end{equation}
The full system state is then formed by stacking the individual states of each of the agents in the set $\mathcal{M}=\{0,1,...,m\}$,
\begin{equation} \label{eq::state}
    \mathbf{q} = \begin{bmatrix}
    \mathbf{q}_0^\top &
    \mathbf{q}_1^\top &
    \hdots &
    \mathbf{q}_m^\top
    \end{bmatrix}^\top.
\end{equation}

The ASV we use is modeled as a nonlinear second order system~\cite{wang_design_2018,wang_roboat_2019,wang_roboat_2020}. Since the vessel sails at low speeds, we discard Coriolis and centripetal effects. Therefore,
\begin{equation}
\begin{aligned}
    \dot{\mathbf{x}}_{i} &= \mathbf{R}\left(\phi_i\right)
    \mathbf{v}_i,\\
    \dot{\mathbf{v}}_{i} &= \mathbf{M}^{-1}\left(\mathbf{B}\mathbf{u}_i - \mathbf{D}\left(\mathbf{v}_i\right) \mathbf{v}_i\right),
\end{aligned}
\end{equation}
where $\mathbf{R}\left(\phi_i\right)$ is the rotation matrix from body to inertial frame, $\mathbf{M}$ is the mass matrix, $\mathbf{B}\mathbf{u}_i$ are the applied forces (thrust) and $\mathbf{D}\left(\mathbf{v}_i\right)$ is the drag matrix.

\subsection{Global Planning} \label{sec:globalplan}
To help navigate large maps, we provide the ego-agent with a global path. Such a path is generated via the ROS navigation stack with its path planning plugin \cite{marder-eppstein_office_2010}. Instead of directly tracking this global path, we look for a local goal $\mathbf{p}_g$ on the global path at a given look-ahead distance $r_{\textbf{p}_g}$ (Fig.~\ref{img:local_goals}). Compared to just rigorously tracking the global path, this local goal approach gives the local planner more freedom to perform collision avoidance and other maneuvers.

%% file: tex/proposed_approach.tex
\section{Decentralized Interaction-aware Model Predictive Path Integral Control}
In the following we outline the proposed architecture, state the changes to the classic MPPI framework and present the regulation-aware cost function along with the local goal prediction used for the decentralized computation of the cost.

\begin{figure}[t]
    \centering
    \medskip
    \medskip
    \includegraphics[width=0.4\textwidth]{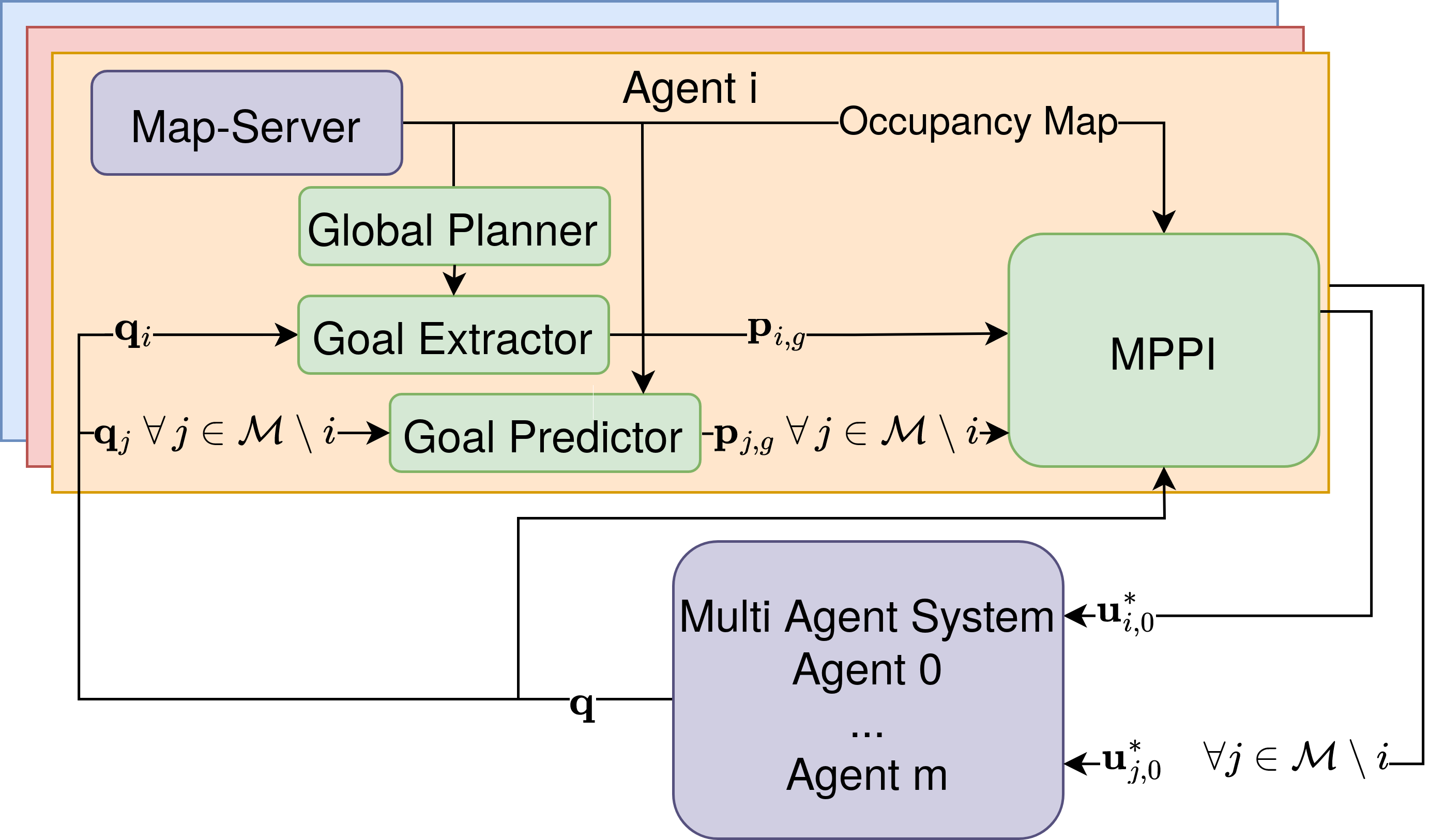}
    \caption{Overview of the framework: At every time step, each agent receives the state of all the agents in the environment. With this, we extract a local goal for the ego-agent (agent $i$ in figure) from the global planner and predict local goals for the other agents. We then solve the planning task with MPPI as if it was centralized, planning a control input sequence and corresponding trajectory for each agent. We then apply the first input $u_{i,0}^*$ of the sequence to the ego-agent.}
    \label{img:overview}
\end{figure}

\subsection{Approach and Architecture}
\begin{figure}[t]
    \centering
    \medskip
    \includegraphics[width=0.45\textwidth]{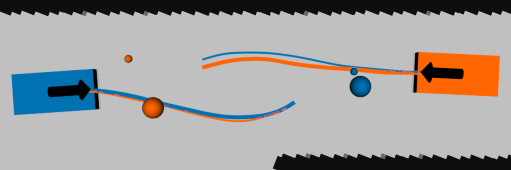}
    \caption{Screenshot of two agents running our decentralized method with no communication inside our simulator. The left (\textcolor{blue}{blue}) agent is given a local goal (\textcolor{blue}{large blue ball}), its state, and the state of the other (\textcolor{orange}{orange}) agent, based on which it predicts a local goal (\textcolor{orange}{small orange ball}) for the obstacle vessel. The \textcolor{blue}{blue} agent plans a sequence of inputs for both itself and the obstacle agent resulting in two trajectories, depicted by the \textcolor{blue}{blue} paths. In \textcolor{orange}{orange}, we can see the other agent applying the same algorithm. Note that even though we use constant velocity to predict the goal of the obstacles, both agents plan cooperation in the collision avoidance.}
    \label{img:optimization_example}
\end{figure}

Our decentralized MPPI approach relies on each agent running its own MPPI solver for its local multi-agent system to anticipate the actions of other agents (see Fig.~\ref{img:overview}). That is, for agent $i$, the MPPI state and control output are defined as,
\begin{equation}
\begin{aligned}
    \mathbf{q}^i&=\begin{bmatrix}\mathbf{q}_i^\top& \mathbf{q}^{i\top}_j\end{bmatrix}^\top,\\
    \mathbf{u}^i&=\begin{bmatrix}\mathbf{u}_i^\top& \mathbf{u}^{i\top}_j\end{bmatrix}^\top,
\end{aligned}
\quad\forall j\in\mathcal{M}\setminus i,
\end{equation}
where $\left(.\right)^i_j$ signifies a variable that agent $i$ estimates of agent $j$. In the centralized case described in Section~\ref{sec:mppi_basic}, the system state is fully observable and the inputs computed by the central controller will be those executed by each agent.
When we move to the decentralized case, the positions and velocities of other agents must be communicated or observed. Furthermore, while each agent samples control actions for all other agents in the MPPI rollouts, at execution time, there is no guarantee that other agents will behave accordingly.
To focus on the decentralized coordination problem, we make a few assumptions. First, we assume noise-free observations of the positions and velocities of other agents, i.e. $\mathbf{q}^i_j=\mathbf{q}_j$. Second, we assume that all agents behave rationally and that they are minimizing the same global cost. We later show in our experiments that the controller is still able to perform well when this assumption is violated. Third, in our experiments we only consider scenarios with homogeneous agents, meaning that they all have the same dynamics. This third assumption is not required in general as considering different dynamical models for different agents is possible as long as models are known. Fig.~\ref{img:optimization_example} shows a simulated encounter between two ASVs running our decentralized algorithm without communication. 

\subsection{Two-stage Sample Evaluation}
With an increasing number of agents, it is increasingly likely for at least one agent to collide with a static obstacle in most rollouts. In the classical implementation of MPPI, this leads to most rollouts receiving a high cost and being effectively rejected, resulting in a very low sample efficiency. Therefore we propose to decouple the sampling into two stages as shown in Algorithm~\ref{alg:mppi_adapted_short}. Control-samples $U_{j,k}$ are evaluated in parallel for every agent $j\in\mathcal{M}$, predicting the set of individual trajectories $Q_{j,k}$ with agent-centric costs $S_{j,k}$ (eq.~\ref{eq::agent_cost}). At this point all samples with cost larger than the collision penalty $C_{\text{collision}}$ are immediately discarded (lines 3 and 4 of Algorithm~\ref{alg:mppi_adapted_short}). To build the expected number of system samples $K$ we sample uniformly from the remaining non-colliding samples of each agent and unify these into full system samples, i.e. by stacking as in eq.~\eqref{eq::state}. For each system sample $Q_k$ the complete configuration cost $S_k$ is evaluated by adding the stored agent-centric costs $S_{j,k}$ with any costs arising from collisions between vessels and regulation violations (line 14, Algorithm~\ref{alg:mppi_adapted_short}).

\begin{algorithm}[th]

\caption{Decentralized MPPI for agent $i$ (agent-specific superscripts are dropped for clarity)} \label{alg:mppi_adapted_short}

\begin{algorithmic}[1]
\Require $U$ \Comment{previous control sequence (hot start)}
\Require $\mathbf{q}$\Comment{current system state}
\State $\mathbf{p}_{i,g} \leftarrow $ \textit{receiveEgoLocalGoal}()
\State $\mathbf{p}_{j,g} \leftarrow $ \textit{predictLocalGoal}$(\mathbf{q}_j)$\Comment{$\forall \,j\in\mathcal{M}\setminus i$}
\For {each agent $j\in\mathcal{M}$}\Comment{independent rollouts}
    \For{each sample $k$}
        \State $\mathcal{E}_{j,k} = \left[\epsilon_{0}, \, \dots, \, \epsilon_{T-1}\right]$\Comment{$\quad \epsilon_t \in \mathcal{N}(0, \nu \mathbf{\Sigma})$}
        \State $\tilde{U}_{j,k} = U_j + \mathcal{E}_{j,k}$
        \State $Q_{j,k} \leftarrow$ \textit{simulateSystem}$(\mathbf{q}_j,\tilde{U}_{j,k})$
        \State  $S_{j,k} \leftarrow$ \textit{getIndividualCost}$(Q_{j,k},\tilde{U}_{j,k},\mathbf{p}_{j,g})$
        \If{$S_{j,k} > C_{\text{collision}}$}
            \State \textit{discardSample}$(Q_{j,k},\tilde{U}_{j,k},\mathcal{E}_{j,k},S_{j,k})$
        \EndIf
    \EndFor
\EndFor
\State Uniformly sample from the remaining valid input sequences to rebuild $K$ full system samples
\For{$k=1:K$}
    \State $\mathcal{S}_k = \sum_{j\in\mathcal{M}}S_{j,k} +         \textit{getConfigurationCost}(Q_{k})$
\EndFor

\State $\big[w_1, \, \dots, \, w_K\big]$ = \textit{importanceSampling}$(\big[\mathcal{S}_0, \, \dots, \, \mathcal{S}_K\big])$
\State return $U^{*} = U + \sum_{k = 1}^{K} w_k \mathcal{E}_k$
\end{algorithmic}
\end{algorithm}

\subsection{Cost Formulation}
\begin{figure}[t]
    \centering
    \medskip
    \includegraphics[width=0.3\textwidth]{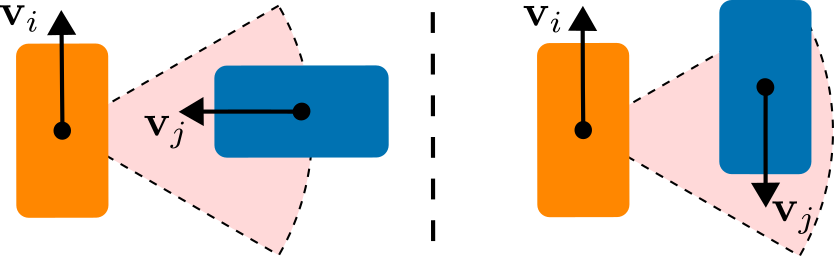}
     \caption{Configurations considered as regulation violations. Left: Not giving right of way to a vessel approaching from starboard. Right: Avoiding an oncoming vessel to the left.}
    \label{img:regulations}
\end{figure}

The sample cost $S_k, \forall k \in [1,K]$  evaluation is split into agent-centric and configuration costs. Both are considered instantaneous costs and are evaluated for every time-step $t$ within the horizon $T$ and each sample $k$. The \textbf{agent-centric} cost $S_{j,k,t}$ (in the following $S_{\text{agent}}$) is evaluated for agent $j$ for sample $k$ and defined as,
\begin{equation}\label{eq::agent_cost}
    S_{\text{agent}} = C_{\text{static}} + C_{\text{rotation}} + C_{\text{tracking}} + C_{\text{speed}} + C_{\text{sample}}.
\end{equation}
$C_{\text{static}}$ returns a constant penalty $C_{\text{collision}}$ if the vessel enters occupied space and $C_{\text{rotation}}$ is based on a linear penalty for rotation velocities ($k_{\text{rot, slow}}$ for velocities $\vert \vert\mathbf{v}\vert \vert_2 < 0.5 \text{m/s}$, $k_{\text{rot}}$ otherwise).
The tracking cost is,
\begin{equation}
    C_{\text{tracking}} = k_{\text{tracking}} \frac{\vert\vert\mathbf{p}_{g} - \mathbf{p}_{t}\vert \vert_2}
    {\vert\vert\mathbf{p}_{g} - \mathbf{p}_{t_0}\vert \vert_2},
\end{equation}
where $k_{\text{tracking}}$ is a scaling factor, $\textbf{p}_g$ is the agent's local goal, $\textbf{p}_{t}$ is the predicted agent position at time $t$ and $\textbf{p}_{t_0}$ is the vessel position at the start of the prediction horizon.
$C_{\text{speed}}$ is a constant penalty applied when the current speed is higher than the maximum speed.
The sample cost is given by,
\begin{equation}\label{eq::sample_cost}
    C_{\text{sample}}(\mathbf{u}_{t}, \mathbf{\epsilon}_{t})\ = \frac{1}{2} \gamma [ \, \mathbf{u}_{t}^T \Sigma^{-1} \mathbf{u}_{t} + 2 \mathbf{u}_{t}^T \Sigma^{-1} \mathbf{\epsilon}_{t} ],
\end{equation}
with $\gamma$ as a tuning parameter.
The \textbf{configuration cost} $S_{k,t}$ (in the following $S_{\text{configuration}}$) is evaluated for every timestep $t$ within the horizon $T$ for every sample $k$ and combines dynamic collisions and regulation violations,
\begin{equation}
    S_{\text{configuration}} = C_{\text{dynamic}} + C_{\text{regulation}}.
\end{equation}
Dynamic collisions are defined as those between multiple vessels and are penalized with the same constant $C_{\text{collision}}$ and the regulation cost $C_{\text{regulation}}$ is derived from the two main traffic rules (i) \emph{avoiding to the right} in head-on encounters and (ii) \emph{right of way} for crossing scenarios similar to COLREGs~\cite{johansen_ship_2016}. Regulation compliance is determined by a relative position and relative velocity check. Regarding the position, we check if there is a vessel with significant velocity ($\vert \vert \mathbf{v} \vert \vert > 0.5 m/s$)  on starboard side within a given radius (Fig~\ref{img:regulations}). Regarding the relative velocities, we evaluate if another vessel approaches from the right by,
\begin{equation}\label{eq::row}
    \vert\vert \mathbf{v}_i \times \mathbf{v}_j \vert \vert < \vert\vert \mathbf{v}_i \vert\vert \, \vert\vert \mathbf{v}_j \vert \vert \sin(-\frac{\pi}{2} + \delta),
\end{equation}
where $\delta$ defines the angular margin, and if the other vessel is approaching the ego-vessel head-on via,
\begin{equation}\label{eq::fs}
    \vert\vert \mathbf{v}_i \cdot \mathbf{v}_j \vert\vert < \vert\vert \mathbf{v}_i \vert\vert \cdot \vert \vert \mathbf{v}_j \vert \vert \cos(\pi + \delta).
\end{equation}
Therefore if we detect a vessel on starboard side and the velocities satisfy \eqref{eq::row}, we consider the ego-agent as breaking the right-of-way rule (Fig.~\ref{img:regulations} left). If instead, we detect a vessel on starboard side with opposite velocity, we consider it as passing on the right (Fig.~\ref{img:regulations} right).

            


\subsection{Local Goal Prediction} \label{sec:goalpred}
\begin{figure}[t]
    \centering
    \medskip
    \includegraphics[width=0.4\textwidth]{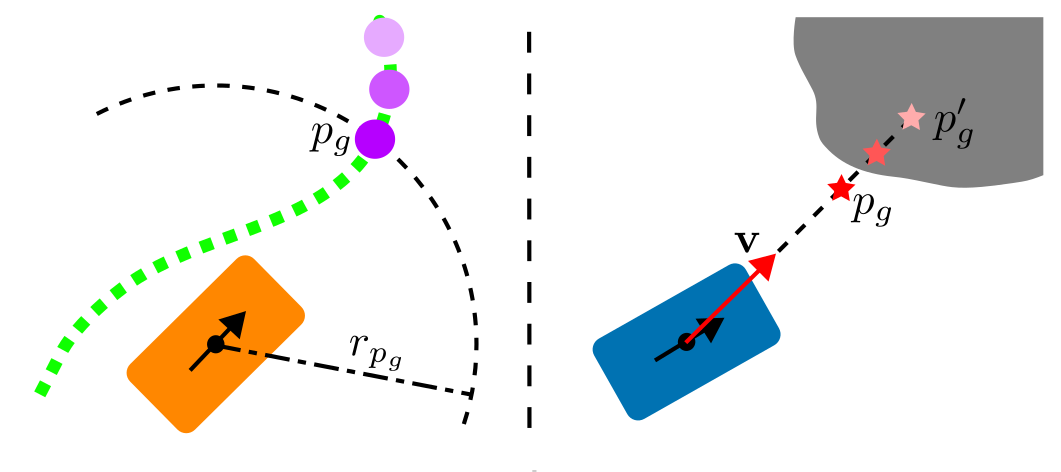}
    \caption{Left: Extraction of the local goal of the ego vessel is performed by searching the global path backward until the goal position is within a radius $r_{p_g}$ from the center of the vessel. Right: Local goal prediction for other agents is performed using a constant velocity model, then projected into unoccupied space if the goal is in collision with static obstacles.}
    \label{img:local_goals}
\end{figure}

The proposed decentralized version of interaction-aware MPPI requires estimating the local goals for all non-ego vessels (line 2, Algorithm~\ref{alg:mppi_adapted_short}). We use a constant velocity model such that agent $i$ estimates the goal of agent $j$ as,
\begin{equation}
    \mathbf{p}^i_{j,g} = k_s \, T \, \delta T \, \mathbf{R}\left(\phi^i_j\right) \mathbf{v}^i_j  + \mathbf{p}^i_{j},
\end{equation}
with $k_s$ as a scaling factor and $\delta T$ as step size. If the predicted goal is in collision, we project back along the vector towards $\mathbf{p}^i_j$ and choose the first unoccupied point as shown in Fig.~\ref{img:local_goals}.

%
%
%

%% file: tex/experiments.tex
\section{Experiments}
We perform extensive experiments in several maps taken from real canal sections of Amsterdam, namely the Herengracht (HG), Prinsengracht (PG), Bloemgracht (BG), and the intersection between Bloemgracht and Lijnbaansgracht (BGLG). In all experiments, the dimensions of both the map and the vessel are represented faithfully.
In Section~\ref{sec::comparison} we compare our method in two-agent scenarios with an optimization-based MPC, in Section~\ref{sec::crowded_scenario} we test our method in interaction-rich four-agent scenarios, in Section~\ref{sec::human_driven} we demonstrate the robustness to non-rational human-driven agents, while in Section~\ref{sec::computational_compl} we discuss the computation times of the proposed method.
All experiments running MPPI use a horizon $T$ of 100 time-steps with step-size $\delta T = 0.1s$, input variance $\Sigma = \text{diag}(0.5,0.5,0.01,0.01)$ and exploration scaling factor $\nu = 12$. We use $K=\{2000,6000\}$ samples for two- and four-agent scenarios, respectively.
Each version of the MPPI shown in the experiments (centralized, decentralized, decentralized with no communications) uses our proposed two-stage sampling technique. 

\begin{figure}[t]
    \centering
    \begin{subfigure}[t]{\linewidth}
    \medskip
    \centering
         \includegraphics[width=0.9\textwidth]{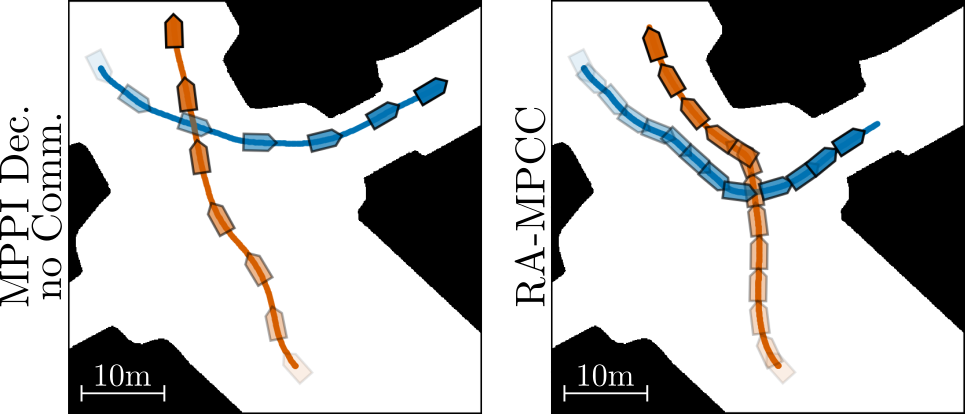}
         \caption{\textit{Prinsengracht}. MPPI understands that the blue vessel can safely cross in front of the vessel with the right of way (orange) without slowing it down. RA-MPCC is not confident and ends up blocking the way, pushing the orange vessel out of its route.}
         \label{fig:compare_prins}
    \end{subfigure}
    \begin{subfigure}[t]{\linewidth}
    \medskip
    \centering
        \includegraphics[width=0.9\textwidth]{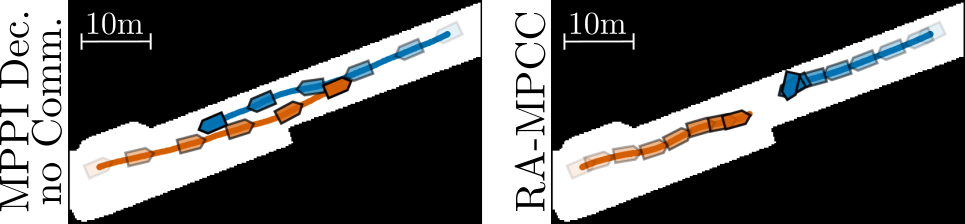}
         \caption{\textit{Bloemgracht}. The MPPI agents cooperate to perform collision avoidance, while RA-MPCC ends up in a deadlock.}
         \label{fig:compare_headon}
    \end{subfigure}
    \begin{subfigure}[t]{\linewidth}
    \medskip
    \centering
        \includegraphics[width=0.9\textwidth]{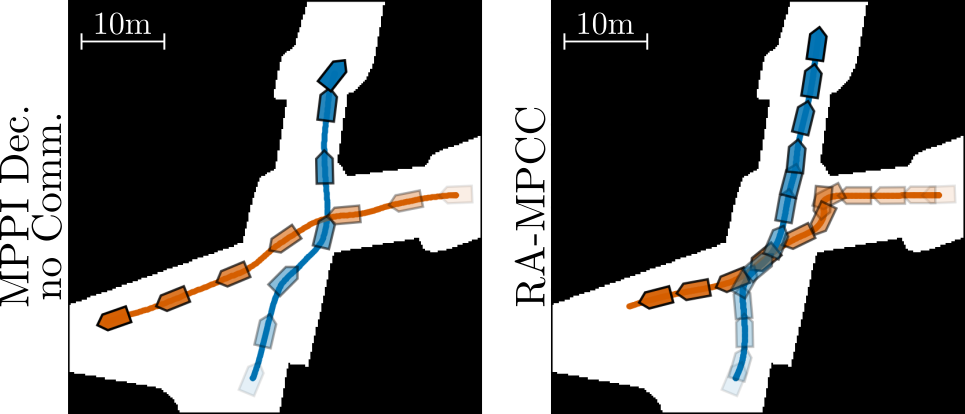}
         \caption{\textit{Bloemgracht-Lijnbaansgracht}. The MPPI agents correctly solve the crossing according to the right of way. RA-MPCC, not understanding interactions, deems much of the state-space as occupied therefore steering left and right. Eventually, the agent with the right of way has to stop and pass behind, violating the navigation rules.}
         \label{fig:compare_crossing}
    \end{subfigure}
    \caption{Comparisons between the decentralized MPPI without communications and RA-MPCC. Vessels on the figures are to scale (4m long) and are plotted every 5 seconds. Results for 100 runs are summarized in Table~\ref{tab::results}.}
    \label{fig:comparisonsimulation}
\end{figure}

\begin{figure*}[ht]
    \centering
    \medskip
    \includegraphics[width=0.7\textwidth]{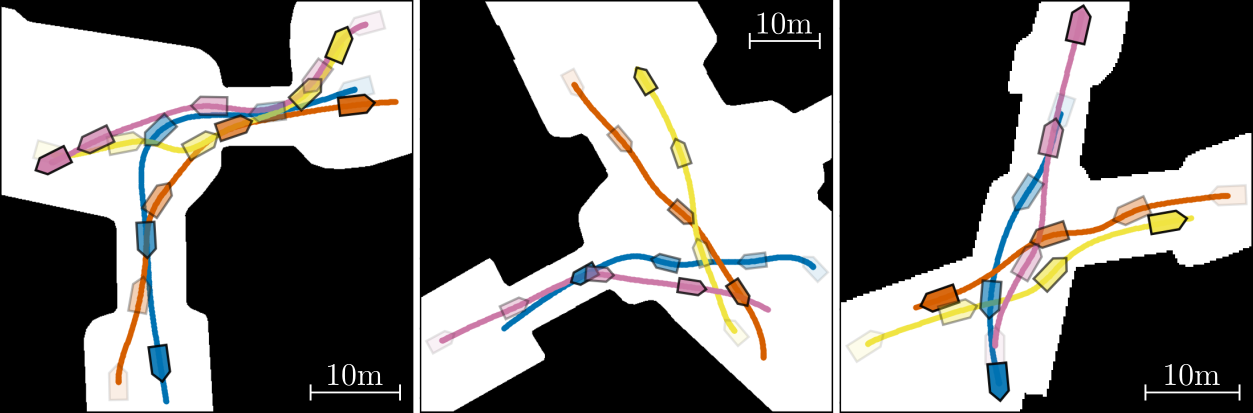}
    \caption{Navigation among four autonomous vessels running decentralized MPPI without communication. Left: Narrow intersection with a bridge in Herengracht. Middle: Wide intersection with bridges in Prinsengracht. Right: Very narrow intersection in Bloemgracht. Vessels are plotted every 6 seconds.}\vspace{-0.5cm}
    \label{img:trajectories}
\end{figure*}

\subsection{Comparison with Optimization-based MPC}\label{sec::comparison} 
We compare our method to a state-of-the-art optimization-based decentralized motion planner designed for ASVs in urban canals, namely the Regulations Aware Model Predictive Contouring Controller (RA-MPCC)~\cite{de_vries_regulations_2022}. The three scenarios on which we compare are an unprotected left turn (Fig.~\ref{fig:compare_prins}), a head-on encounter (Fig.~\ref{fig:compare_headon}) and a crossing (Fig.~\ref{fig:compare_crossing}). These three scenarios were then run 100 times with randomized initial conditions and global goals using the proposed centralized, decentralized, and decentralized with no communication MPPI as well as the RA-MPCC. For all controller types, the randomization was kept equal (i.e. same random seed). Table~\ref{tab::results} summarizes the results, where we compare the number of runs that ended successfully, in a deadlock, or in a collision. Of the runs that ended successfully, we report the number of rule violations. We also report the average time to complete the scenario, defined as the moment in which all agents reach their goal, and the total average distance, which is the sum of the average distances traveled by all agents.

All controllers were encouraged through the cost function to keep a velocity around the speed limit (around 1.7m/s). From the table, however, it stands out that the RA-MPCC navigates much slower and therefore has much longer arrival times. This is both because of how its cost function is defined, but also because the RA-MPCC has to plan within convex obstacle-free areas, which can sometimes be quite small and slow down the pace. Instead, MPPI considers the exact occupancy map without any need for pre-processing.

While it is easy to give a discontinuous penalty to the MPPI whenever a sample violates a navigation rule, the RA-MPCC has to use continuous cost functions to encourage rule compliance. This, however, inadvertently introduces repulsive forces between the two agents, which then tend to push each other into corners, which is the main cause of deadlocks in the Prinsengracht and the Bloemgracht-Lijnbaansgracht scenarios.
In the Bloemgracht head-on encounter, the RA-MPCC gets to its destination only about half of the time. Given that the RA-MPCC has to inflate the ego-agent in a set of circles, the obstacle agent into an ellipsoid, and the static obstacle map has to be pre-processed into convex regions, there is barely enough space to pass in the most narrow section of the canal. This, combined with the lack of understanding that the two agents can cooperate to solve the maneuver, leads to a large number of deadlocks.

Collisions with the RA-MPCC instead happen for two reasons. Number one, the method first approximates the static obstacle with polygons, which are then decomposed into convex shapes, to which we can then find linear constraints by solving a quadratic program. However, there is no guarantee that the polygons contain all of the original obstacle. Safety margins are added, but margins too large means that some narrow canals are simply impossible to navigate. Number two, the optimization can often just fail, especially in more difficult and risky situations. When this happens, the algorithm just applies zero input, and if the boat has enough momentum it can drift into a collision.

Moreover, while our interaction-aware MPPI could plan with a horizon of 100 steps, the RA-MPPC could only plan 20 steps to meet the 10Hz control loop. 

\begin{table}[h]
\begin{center}
\caption{Results for 100 runs of the experiments seen in Fig.~\ref{fig:comparisonsimulation} with randomized initial conditions and goals.} \label{tab::results}
\setlength{\tabcolsep}{3pt}
\begin{tabular}{c l c c c c}
 \toprule
 & \multirow{3}{*}{\textbf{Method}} & \textbf{Successes-} & \multirow{2}{*}{\textbf{Rule}} & \multirow{2}{*}{\textbf{Average}}  & \textbf{Total}\\
 &                                  & \textbf{Deadlocks-} & \multirow{2}{*}{\textbf{Violations}} & \multirow{2}{*}{\textbf{Time}}   & \textbf{Average}\\
 &                                  & \textbf{Collisions} &                                      &    & \textbf{Distance}\\
 \midrule
\parbox[t]{2mm}{\multirow{4}{*}{\rotatebox[origin=c]{90}{PG}}} 
& Centralized  &  100 - 0 - 0  &  2  &  24.65s  &  82.15m \\
& Dec. Comm.  &  100 - 0 - 0  &  2  &  24.64s  &  82.14m \\
& Dec. No Comm.  &  100 - 0 - 0  &  2  &  25.12s  &  82.54m \\
& RA-MPCC & 95 - 5 - 0 & 5 & 51.81s & 73.20m\\
\midrule
\parbox[t]{2mm}{\multirow{4}{*}{\rotatebox[origin=c]{90}{BG}}}
& Centralized  &  100 - 0 - 0  & 0 &  22.23s  &  74.93m \\
& Dec. Comm.  &  100 - 0 - 0  & 0 &  22.24s  &  74.79m \\
& Dec. No Comm.  &  100 - 0 - 0  & 0 &  22.18s  &  75.01m \\
& RA-MPCC     &   52 - 37 - 11 & 4 &  55.93s & 67.64m \\
\midrule
\parbox[t]{2mm}{\multirow{4}{*}{\rotatebox[origin=c]{90}{BGLG}}}
& Centralized  &  100 - 0 - 0  &  2  &  28.47s  &  87.24m \\
& Dec. Comm.  &  100 - 0 - 0  &  2  &  28.83s  &  88.26m \\
& Dec. No Comm.  &  100 - 0 - 0  &  2  &  29.01s  &  88.30m \\
& RA-MPCC & 74 - 23 - 3 & 38 & 54.46s & 84.20m\\

 \bottomrule
\end{tabular}
\end{center}
\end{table}

\subsection{Navigation in Crowded Environments}\label{sec::crowded_scenario}
The proposed method is also capable of resolving scenarios with more than two agents. In Table~\ref{tab::results4} we show the results for 20 runs in crowded environments with four agents. The experiments are run in the maps shown in Fig.~\ref{img:trajectories}, where the vessels are exposed to many encounters in very narrow spaces. The results show that the centralized and decentralized method with shared local goals (with communication) can resolve the task successfully while behaving cooperatively. 
The decentralized version of our approach with no communication (thus predicting goals for other vessels) incurs in a collision in the tight Bloemgracht's intersection. With the four agents so close to each other, the vessels need to perform large avoidance maneuvers. This causes the estimated local goals and corresponding predictions to diverge drastically from the ground truth. 
However, similarly to the results in Table~\ref{tab::results}, decentralization has little effect on arrival time, total distance, and rule violations.


\begin{table}[t]
\begin{center}
\caption{Results for 20 runs of the experiments seen in Fig.~\ref{img:trajectories} with randomized initial conditions and goals.} \label{tab::results4}
\setlength{\tabcolsep}{3pt}
\begin{tabular}{c l c c c c c}
\toprule
 & \multirow{3}{*}{\textbf{Method}} & \textbf{Successes-} & \multirow{2}{*}{\textbf{Rule}} & \multirow{2}{*}{\textbf{Average}}  & \textbf{Total}\\
 &                                  & \textbf{Deadlocks-} & \multirow{2}{*}{\textbf{Violations}} & \multirow{2}{*}{\textbf{Time}}   & \textbf{Average}\\
 &                                  & \textbf{Collisions} &                                      &    & \textbf{Distance}\\

\midrule
\parbox[t]{2mm}{\multirow{3}{*}{\rotatebox[origin=c]{90}{HG}}} 
 & Centralized & 20 - 0 - 0  & 0  & 24.46  & 204.42 \\
 & Dec. Comm. & 20 - 0 - 0  & 0  & 26.82  & 214.64 \\
 & Dec. No Comm. & 20 - 0 - 0  & 3  & 33.18  & 225.48 \\
  
\midrule
\parbox[t]{2mm}{\multirow{3}{*}{\rotatebox[origin=c]{90}{PG}}} 
 & Centralized & 20 - 0 - 0  & 0  & 23.24  & 214.92 \\
 & Dec. Comm. & 20 - 0 - 0  & 0  & 23.82  & 216.99 \\
 & Dec. No Comm. & 20 - 0 - 0  & 0  & 23.43  & 214.71 \\
  
\midrule
\parbox[t]{2mm}{\multirow{3}{*}{\rotatebox[origin=c]{90}{BGLG}}} 
 & Centralized & 20 - 0 - 0  & 1  & 18.36  & 143.14 \\
 & Dec. Comm. & 20 - 0 - 0  & 2  & 20.87  & 149.78 \\
 & Dec. No Comm. & 19 - 0 - 1  & 0  & 20.13  & 144.86 \\
\bottomrule

\end{tabular}   
\end{center}
\end{table}



\subsection{Navigation among Human-piloted Vessels}\label{sec::human_driven}
\begin{figure}[t!]      
    \centering
    \medskip
    \includegraphics[width=0.4\textwidth]{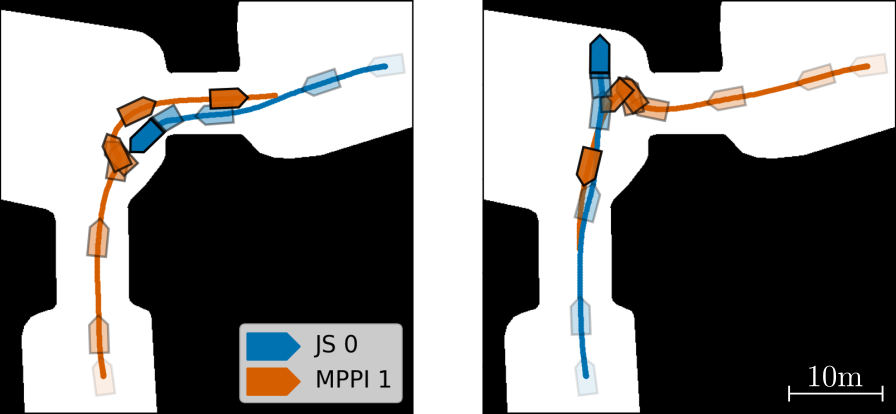}
    \caption{Qualitative robustness evaluation for non-cooperative vessels. Left: The joystick-driven vessel (blue) avoids to the wrong side. The MPPI vessel (orange) avoids collisions by coming to a complete stop, then continues to the left. Right: The joystick-driven vessel (blue) blocks the MPPI, disregarding right of way. The MPPI agent (orange) avoids collisions, comes to a stop, then continues on its way. Vessels are plotted every 4 seconds.}
    \label{img:human_trajectories}
\end{figure}

As long as the human-piloted agent behaves rationally, the resulting trajectories are very similar to the ones presented in Fig.~\ref{fig:comparisonsimulation}, where all agents are autonomous. Therefore, in Fig.~\ref{img:human_trajectories} we demonstrate how the proposed decentralized MPPI with no communication can cope with irrational agents.


\subsection{Computational Complexity}\label{sec::computational_compl}
As previously found in the literature~\cite{williams_model_2017},  we confirm that MPPI scales linearly with an increasing number of agents (with constant sample number $K$). Table~\ref{tab::comp_time} shows that the algorithm runs at about 10Hz with two agents, therefore in real-time, and down to less than 4Hz with five agents. We want to stress that this was achieved with parallelization of the sampling procedure over the CPU (Intel® Xeon® W-2123 CPU @ 3.60GHz × 8, 64 GB). Parallelizing this algorithm on a GPU would be highly beneficial, allowing for real-time control of several agents~\cite{williams_model_2017}. 

\begin{table}[h]
\begin{center}
\caption{Average computation time $t_c$ and standard deviation $\sigma_{t,c}$ for increasing number of agents} \label{tab::comp_time}
\setlength{\tabcolsep}{3pt}
\begin{tabular}{c c c c c }
 \toprule
 \textbf{Number of agents} & \textbf{2} & \textbf{3} & \textbf{4} & \textbf{5} \\
 \midrule
$t_c$ (ms)  & 90.7 & 137.2 & 182.9 & 209.9  \\
$\sigma_{t_c}$ (ms)  & 6.7 & 8.9 & 17.5 & 26.0 \\

\bottomrule

\end{tabular}
\end{center}
\end{table}

%% file: tex/conclusions.tex
\section{Conclusions}
\label{sec:conclusions}
Within this work, we developed an MPPI controller for decentralized interaction-aware navigation in urban canals. In multiple sets of randomized scenarios, we demonstrate that our method outperforms a state-of-the-art MPC in terms of success rate, deadlocks, collisions, rule violations, and arrival times. In extensive experiments among several rational autonomous agents and case studies with potentially non-cooperative human drivers, we show robust operation while providing insights into the limitations of the approach. We display that decentralizing the MPPI does not sacrifice performance. Moreover, we demonstrate that the method would be able to run in real-time with multiple agents. In the future, however, a GPU implementation would vastly reduce the computation time. Moreover, the sampling distribution can be improved, thus requiring fewer samples to obtain a good approximation of the optimal control.

